\documentclass[preprint,12pt]{elsarticle}
\usepackage{amssymb}
\usepackage{amsmath}
\usepackage{graphicx}
\usepackage{float}
\usepackage{booktabs}
\usepackage{tabularx} 
\usepackage{xcolor}
\usepackage{algorithm}
\usepackage{algorithmic}
\usepackage{caption}
\usepackage{url} 
\usepackage{enumitem}
\usepackage{comment}

\newcommand{\best}[1]{\textcolor{red}{#1}}
\newcommand{\second}[1]{\textcolor{blue}{#1}}
\newcommand{\third}[1]{\textcolor{green!60!black}{#1}}

\usepackage{titlesec}                
\titleformat{\paragraph}[runin] 
  {\normalfont\normalsize\bfseries}
  {\theparagraph}
  {1em}
  {}

\begin{document}

\begin{frontmatter}
 \title{Towards Modality-Agnostic Medical Image Anomaly Detection: A Training-Free Manifold Refinement Approach}

\author[inst1]{Pritam Kar}
\ead{karpritam25@iisertvm.ac.in}

\author[inst1]{Gouri Lakshmi S}
\ead{gourilakshmis21@iisertvm.ac.in}

\author[inst1]{Saptarshi Bej\corref{cor1}}
\ead{sbej7042@iisertvm.ac.in}

\affiliation[inst1]{
  organization={School of Data Science, Indian Institute of Science Education and Research},
  city={Thiruvananthapuram},
  postcode={695551},
  state={Kerala},
  country={India}
}

\cortext[cor1]{Corresponding author}

\begin{abstract}
Deploying AI-based anomaly detection across diverse clinical imaging settings remains challenging because most existing methods rely on modality-specific architectures, anatomical priors, or extensive retraining, limiting their use as general-purpose screening tools in heterogeneous healthcare environments. One-class classification (OCC) offers a label-efficient alternative by training exclusively on normal data, but conventional two-stage pipelines fit a density estimator directly on raw pretrained embeddings, leaving substantial discriminative structure in the latent space unexploited. We introduce a training-free, modality-agnostic framework that inserts an explicit manifold-refinement stage between feature extraction and anomaly scoring. Empirical density weights, estimated via a UMAP-derived neighborhood graph, guide an iterative shift of embeddings toward locally dense regions, compacting normal samples while leaving anomalies relatively isolated-prior to Gaussian density estimation and Mahalanobis-based scoring. Critically, this refinement introduces no additional trainable parameters and requires no architectural modification, allowing it to be layered onto any pretrained encoder. Evaluated on the MedIAnomaly benchmark across seven datasets spanning five imaging modalities (chest X-ray, brain MRI, retinal fundus, dermatoscopy, histopathology), the framework achieves the best AUC on four datasets and the best Average Precision on five datasets among methods evaluated in the MedIAnomaly benchmark, outperforming specialized reconstruction and diffusion-based methods using a single fixed hyperparameter configuration across all modalities. These results demonstrate that meaningful gains in anomaly detection can be achieved through post-hoc geometric refinement of existing representations rather than through bespoke encoders, offering a practical and scalable AI screening framework for real-world, multi-modality clinical workflows where retraining and abnormal-case annotation are costly or infeasible.
\end{abstract}

\begin{keyword}
Anomaly Detection \sep
Medical Image \sep
Manifold Learning \sep
Density Enhancement \sep
One-Class Classification \sep
Modality Agnostic
\end{keyword}

\end{frontmatter}

\section{Introduction}

\textbf{One Class Classification (OCC)-based anomaly detection in medical imaging. }In many clinical settings, obtaining large-scale annotated datasets of abnormal cases is challenging due to the rarity of certain pathologies, the high cost of expert annotation, and privacy constraints~\cite{CAI2025103500, ruff2021unifying}. Consequently, anomaly detection has emerged as a promising alternative to conventional supervised learning. Among anomaly detection paradigms OCC-based methods are particularly attractive because they learn exclusively from normal data and identifies deviations from the learned normal distribution as potential abnormalities~\cite{ruff2021unifying, CAI2025103500}. This enables the detection of rare, diverse, or previously unseen pathologies without requiring labeled abnormal examples during training, making OCC a scalable and label-efficient framework for early disease screening and clinical decision support~\cite{CAI2025103500, bercea_towards_2024}. As a result, a large body of research has focused on developing increasingly sophisticated OCC-based anomaly detection methods for medical imaging.

\textbf{Modality-agnostic anomaly detection and the role of the OCC paradigm in it. }Many recent anomaly detection methods achieve impressive performance by leveraging modality-specific architectures, anatomical priors, or training strategies tailored to particular imaging domains~\cite{bercea_towards_2024, lagogiannis2023unsupervised, CAI2025103500}. Such specialization is often beneficial, as it enables models to capture fine-grained characteristics of specific pathologies and imaging modalities. For example, diffusion-based methods such as AutoDDPM excel on histopathological textures~\cite{bercea2023autoddpm}, while reconstruction-based approaches such as AE-PL achieve strong performance in retinal imaging~\cite{CAI2025103500}. However, the same specialization that drives high performance can limit rapid deployment across heterogeneous clinical settings, where a single screening tool must operate reliably across multiple imaging modalities and disease types without extensive retraining, architectural redesign, or large quantities of annotated abnormal data.

In practice, large healthcare systems frequently require an initial screening tool capable of operating across diverse modalities and pathology types. Existing methods are often optimized for specific datasets, pathologies, or modalities, making their direct transfer to new clinical environments challenging~\cite{bercea_towards_2024}. Addressing this challenge requires anomaly detection frameworks that are both modality-agnostic and inherently capable of identifying previously unseen abnormalities. In this context, the OCC paradigm provides a natural foundation, as it learns a compact representation of normality and flags deviations from it, eliminating the need for pathology-specific supervision while retaining the ability to generalize across diverse clinical scenarios. To summarize, the key advantages of the OCC paradigm in the context of anomaly detection in medical imaging are:

\begin{enumerate}[label=(\roman*)]
    \item Eliminating the need for abnormal labels, thereby reducing annotation cost and effort.
    \item Enabling potential modality-agnostic deployment across diverse imaging settings.
    \item Naturally addressing the severe class imbalance inherent in medical datasets.
    \item Facilitating early disease screening and clinical triage by identifying deviations from normality.
\end{enumerate}

\textbf{A brief overview of OCC methods. }Traditional OCC-based anomaly detection approaches in medical imaging can be broadly grouped into reconstruction based, self-supervised learning (SSL)-based, and feature reference-based methods; we discuss each category in detail in Section~\ref{sec:related_work}~\cite{CAI2025103500}. For the purposes of motivating our contribution, it suffices to note that the two-stage SSL paradigm in which a pretext head used to learn discriminative representations from normal data is discarded after training, and a density estimator, most commonly a Gaussian density estimator (GDE), is instead fitted on the resulting embeddings has been shown to be particularly competitive when paired with ImageNet-pretrained feature extraction, outperforming both reconstruction-based and feature reference-based alternatives in medical anomaly detection~\cite{CAI2025103500, li2021cutpaste, sato2023anatomy}. Methods such as CutPaste~\cite{li2021cutpaste} and its anatomy-aware variant AnatPaste~\cite{sato2023anatomy} exemplify this paradigm, learning representations through synthetic transformations applied to normal training data and subsequently scoring anomalies via a GDE fitted on the resulting embeddings.

\textbf{Improving OCC through latent-space density enhancement. }We build directly on the two-stage SSL paradigm and its density-based scoring rule, and show that the resulting performance can be further improved through latent-space density enhancement applied between feature extraction and scoring. Specifically, we propose a framework that combines pretrained feature extraction with a density-enhancement mechanism operating entirely in the embedding space. 

Medical images are first embedded into a high-dimensional latent space using pretrained neural network backbones. The resulting embeddings are then shifted toward regions of higher empirical likelihood using a scalable adaptation of the Mean-Shift algorithm, exposing anomalous instances and improving their separability from normal samples. Intuitively, density enhancement iteratively moves normal embeddings toward local modes of the latent distribution, increasing cluster compactness while causing anomalous embeddings to remain relatively isolated, thereby improving their distinguishability during anomaly scoring.

It is important to note here that, unlike the standard two-stage paradigm, in which the density estimator is fitted directly on the raw extracted embeddings, \textit{our approach performs density enhancement} as an intermediate manifold refinement step prior to anomaly scoring, which is the key novelty of the proposed method. Specifically, density enhancement is achieved through an iterative manifold shifting procedure that moves embeddings toward locally denser regions of the feature space. Throughout the remainder of the paper, we use the terms \emph{density enhancement} and \emph{manifold shifting} interchangeably, with the former emphasizing the objective and the latter referring to the underlying mechanism.

The proposed framework builds upon the OCC paradigm towards a general-purpose anomaly detection framework for rapid screening across diverse imaging modalities. By operating entirely on pre-extracted feature embeddings, the method remains independent of the underlying imaging modality. Furthermore, the manifold refinement and anomaly scoring stages introduce no additional trainable parameters, allowing embeddings generated by pretrained encoders to be processed without retraining, retuning, or architectural modification. Experimental evaluation on the MedIAnomaly benchmark, comprising seven datasets spanning five imaging modalities: chest X-ray, brain MRI, retinal fundus, dermatoscopy, and histopathology, demonstrates the effectiveness and robustness of the proposed approach~\cite{CAI2025103500, medianomaly_repo}. The main contributions of this work are summarized as follows:

\begin{enumerate}[label=(\roman*)]

\item We propose a modality-agnostic anomaly detection framework that operates directly on pretrained feature embeddings, requiring no backbone retraining, architectural modification, or modality-specific tuning, making it practical and scalable across diverse clinical imaging settings.

\item We introduce a scalable Mean-Shift density-enhancement mechanism that refines the latent embedding geometry prior to scoring, improving the separability of normal and anomalous samples without adding any trainable parameters.

\item We formulate a label-free transductive scoring scheme, with the density estimator frozen on normal training data alone, ensuring no abnormal-case annotations are required at any stage.

\item Using a single fixed hyperparameter configuration, we achieve state-of-the-art AUC on four and state-of-the-art Average Precision on five of seven datasets spanning five imaging modalities in the MedIAnomaly benchmark, with a leakage-free tuning study confirming this default remains a robust choice in practice.
\end{enumerate}

\section{Related Work}
\label{sec:related_work}

The proposed framework combines two-stage SSL-based OCC, density enhancement of the latent space, and Gaussian-based anomaly scoring. This section reviews the literature most directly relevant to these three components and identifies the gap that motivates the present work.

\textbf{OCC-based anomaly Detection in Medical Imaging. }Anomaly detection in medical imaging has been extensively studied due to the challenge of identifying rare pathological conditions without comprehensive labeled datasets. Supervised approaches are limited by the scarcity of annotated abnormal samples, motivating the development of OCC and unsupervised methods that train exclusively on normal data~\cite{ruff2021unifying, CAI2025103500}. In literature there are three types of OCC-based methods, namely reconstruction-based methods, self-supervised learning (SSL)-based methods, and feature reference-based methods.

\textit{Reconstruction-based methods}, including autoencoders~\cite{baur2021autoencoder_survey}, variational autoencoders~\cite{kingma2013vae, zimmerer2019vae_anomaly}, and generative adversarial networks~\cite{schlegl2017anogan, schlegl2019fanogan}, learn to reconstruct normal images and flag anomalies through elevated reconstruction error. While effective at capturing global structure, these approaches often struggle with subtle or localised abnormalities, and their performance is sensitive to the choice of reconstruction loss and latent space configuration~\cite{meissen2022pitfalls_reconstruction, CAI2025103500}.

\textit{Self-supervised learning (SSL)-based methods}, particularly those following the two-stage paradigm, address these limitations by first learning discriminative representations via a pretext task on normal data alone, then discarding the pretext head and fitting a density estimator typically a Gaussian density estimator (GDE) on the resulting embeddings~\cite{li2021cutpaste, sato2023anatomy, reiss2021panda}. \textit{Feature reference-based methods} instead compare a test sample's features against reference features obtained via knowledge distillation~\cite{bergmann2020student_teacher} or memory-bank prototypes~\cite{defard2021padim}, typically built on ImageNet-pretrained networks; although popular in industrial defect detection, this paradigm has been shown to transfer poorly to medical imaging~\cite{CAI2025103500}. A comprehensive evaluation by Cai et al.~\cite{CAI2025103500} across seven medical datasets spanning five imaging modalities demonstrates that, among these paradigms, methods leveraging ImageNet-pretrained feature extraction whether used directly, adapted within the two-stage SSL paradigm, or employed for feature reconstruction consistently outperform reconstruction-based and feature reference-based counterparts. Notably, the same study finds that a fixed ImageNet-pretrained ResNet18 used as a feature extractor within the two-stage paradigm, followed by a Gaussian density estimator, achieves competitive or superior performance to more complex specialised methods across modalities. This finding directly motivates our approach: rather than redesigning the backbone or the scoring head, we focus on improving the quality of the feature embeddings themselves prior to scoring.

\subsection{Density-Based Anomaly Scoring within the Two-Stage SSL Paradigm}

Within the two-stage SSL paradigm, anomaly scores are most commonly obtained by modelling the distribution of normal embeddings in feature space and measuring deviations from this distribution, once the pretext head has been discarded~\cite{CAI2025103500}. The Gaussian density estimator (GDE) with Mahalanobis distance scoring~\cite{demaesschalck2000mahalanobis} is the most widely used scoring rule in this context, as it accounts for both the mean and the covariance structure of the normal feature distribution. When applied to representations learned through self-supervised pretext tasks, GDE-based scoring has been shown to be effective across a range of medical imaging benchmarks~\cite{li2021cutpaste, sato2023anatomy, CAI2025103500}.

However, the effectiveness of GDE-based scoring depends critically on the geometric properties of the embedding space: if normal samples do not form a compact, well-separated cluster in feature space, distance-based scores become unreliable. Existing two-stage SSL methods do not explicitly address this geometric precondition and rely on the pretext task alone to induce the necessary structure before the GDE is fitted. The proposed framework explicitly targets this gap by refining the embedding geometry prior to density estimation.

\subsection{Manifold Learning and Mean-Shift Density Enhancement}

Manifold learning techniques aim to capture the intrinsic geometry of high-dimensional data by modelling local neighbourhood relationships. Methods such as Laplacian Eigenmaps~\cite{belkin2003laplacian}, PCA~\cite{jolliffe2016pca}, and UMAP~\cite{mcinnes2018umap} are typically used to project high-dimensional feature distributions onto a lower-dimensional embedding that preserves local neighbourhood structure, and have been applied to anomaly detection to improve the separation between normal and anomalous samples within that reduced space~\cite{ruff2021unifying}. In contrast to this use of manifold learning for \emph{dimensionality reduction}, the present work uses manifold structure purely to \emph{refine} the latent space without changing its dimensionality: embeddings are shifted within their original, full-dimensional feature space toward regions of higher empirical likelihood, rather than being projected onto a lower-dimensional manifold for downstream use.

The Mean-Shift algorithm is a classical nonparametric density estimation technique that iteratively moves each data point toward the local mode of the estimated density. Applied in feature space, Mean-Shift has the effect of compacting clusters of high-density points while leaving low-density outliers relatively isolated, a property that is directly useful for OCC. However, vanilla Mean-Shift relies on fixed bandwidth kernels and does not scale well to high-dimensional embedding spaces, where density estimation becomes unreliable due to the curse of dimensionality.

The present work introduces an empirical-likelihood-weighted variant of Mean-Shift that borrows a single component of UMAP the construction of a fuzzy simplicial set over the $k$-nearest-neighbour graph~\cite{mcinnes2018umap} purely as a tool for estimating local density, rather than for dimensionality reduction or visualisation as in conventional UMAP usage. The resulting fuzzy neighbourhood affinities are used only to compute adaptive, data-driven weights for the shift operation; the embeddings themselves remain in their original feature space throughout the refinement process. This design avoids explicit bandwidth selection and remains tractable at the dimensionalities typical of pretrained image encoders. To our knowledge, no prior work has applied iterative, dimensionality-preserving manifold-based density enhancement as a refinement step between feature extraction and density-based anomaly scoring in medical imaging.

\subsection{Summary and Positioning}

Existing OCC methods for medical image anomaly detection address representation quality, density-based scoring, or manifold geometry in isolation. Two-stage SSL methods improve representations through a pretext task but treat the resulting embedding space as fixed once the GDE is fitted. GDE-based scoring performs well when embeddings are already well-structured but does not act to improve that structure. Manifold learning methods have been applied for visualisation and dimensionality reduction but not as a dimensionality-preserving intermediate refinement step in an anomaly scoring pipeline. The proposed framework is, to the best of our knowledge,  the first to combine these three elements into a unified, training-free, modality-agnostic pipeline: pretrained backbone feature extraction, iterative manifold-based density enhancement of the embedding space at its original dimensionality, and Mahalanobis distance scoring on the refined embeddings.

\section{Methodology}
\label{sec:methodology}

\begin{figure}[t]
    \centering
    \includegraphics[width=\linewidth]{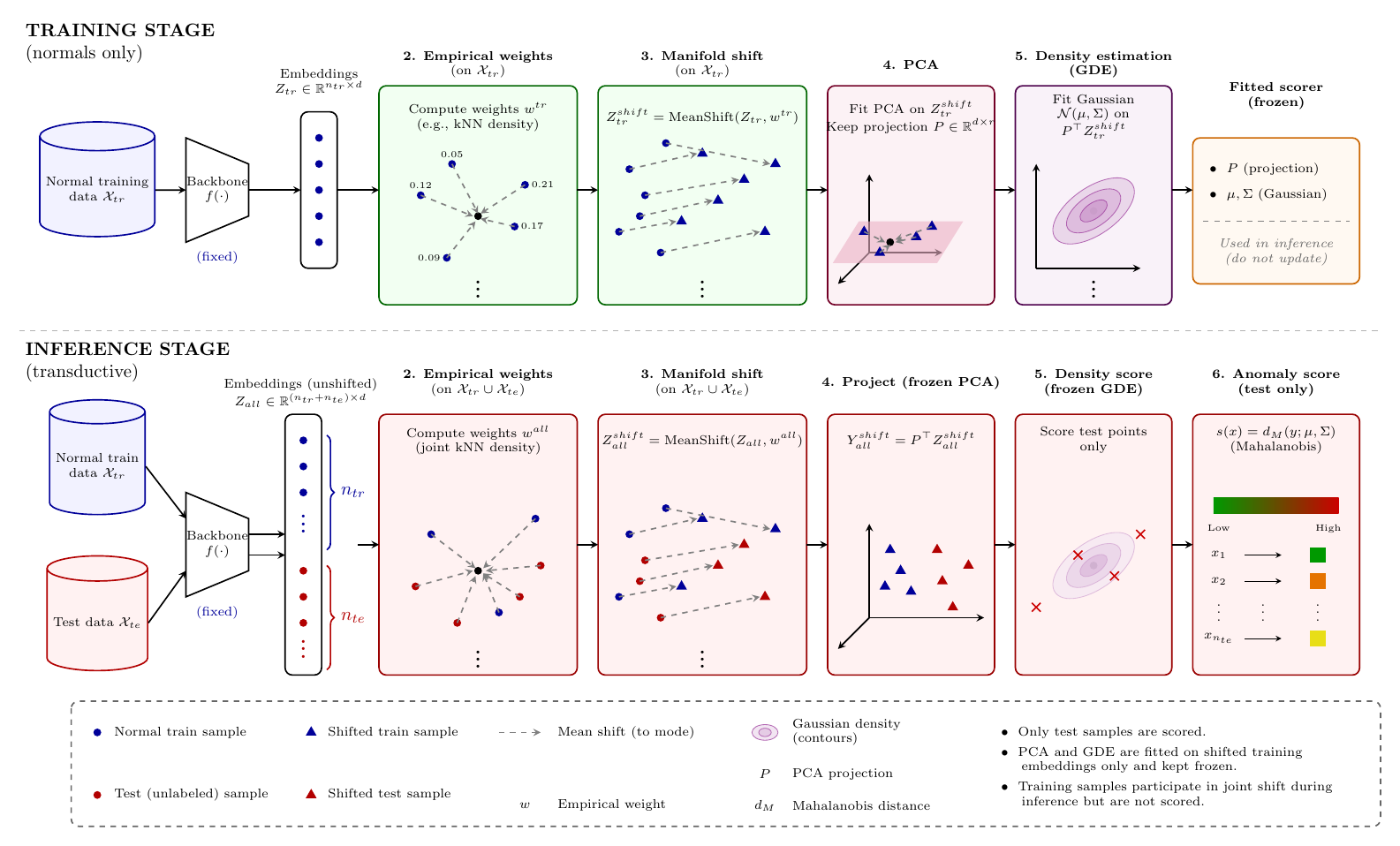}
    \caption{Overview of our pipeline (Algorithm~\ref{alg:msde}). \textbf{Training (normals only):} embeddings from a fixed pretrained backbone (AnatPaste for RSNA, VinDr-CXR, and ISIC2018; ImageNet-pretrained ResNet18 for the remaining datasets) are assigned empirical density weights (\textsc{EmpWeights}), iteratively shifted toward higher-density regions (\textsc{ShiftManifold}), and used to fit a frozen PCA + Gaussian density estimator (GDE). \textbf{Inference (transductive):} weights and shifts are recomputed jointly over train+test embeddings so test points are evaluated in the same manifold context; only the shifted test embeddings are scored via the frozen PCA/GDE, yielding Mahalanobis distances that are standardized and passed through a sigmoid to give calibrated anomaly scores in $(0,1)$, with higher scores indicating greater likelihood of pathology.}
    \label{fig:pipeline}
\end{figure}

The proposed anomaly detection framework consists of two main stages: (i) feature extraction using deep neural network backbones, and (ii) mean shift-based feature refinement, comprising feature-space density enhancement followed by transductive inference and anomaly scoring on the refined embeddings. An overview of the pipeline is illustrated in Fig.~\ref{fig:pipeline}, which depicts both the training stage (top row) and the transductive inference stage (bottom row); in each row, block~1 corresponds to stage~(i), and the remaining blocks correspond to stage~(ii). We describe each stage in detail below.

\subsection{Feature Extraction Using Deep Neural Network Backbones}
The first stage of the proposed pipeline involves extracting discriminative feature representations from medical images using deep convolutional neural networks. Following the experimental setup of MedIAnomaly~\cite{CAI2025103500}, we employ two feature extraction strategies depending on the dataset characteristics. Briefly, we employ either a generic ImageNet-pretrained backbone for broad cross-modality feature extraction or a domain-adapted self-supervised backbone when anatomical priors and modality-specific structure can provide more informative representations. More details can be found in Section \ref{sec:implementation_details}.

Given an input image $I_i$, the feature extraction process is formally expressed as:
\begin{equation}
\mathbf{x}_i = f_{\theta}(I_i),
\end{equation}
where $f_{\theta}(\cdot)$ denotes the feature extraction function parameterized by the ResNet18 encoder (either ImageNet-pretrained or AnatPaste-trained), and $\mathbf{x}_i \in \mathbb{R}^{d}$ is the resulting d-dimensional feature vector. In our experiments we took $d=512$.

Although our methodology is demonstrated here using generic ImageNet-pretrained backbones, it is equally compatible with domain-adapted self-supervised backbones, without requiring any modification to the refinement or scoring stages. In medical imaging, methods such as CutPaste~\cite{li2021cutpaste} synthesise pseudo-anomalies through patch-level augmentations and train the network to distinguish normal from altered samples, with a Gaussian density estimator applied to the resulting embeddings at inference. Domain-specific extensions such as AnatPaste~\cite{sato2023anatomy} incorporate anatomically meaningful transformations for chest radiographs and improve performance on chest X-ray benchmarks. These methods focus primarily on improving the pretext task and treat the downstream embedding space as fixed once training is complete. In this work, we use the pre-extracted embeddings supplied by the MedIAnomaly benchmark for both backbone types (Section~\ref{sec:implementation_details}); we do not perform any additional fine-tuning of either backbone ourselves. To summarize, our framework operates downstream of feature extraction and is agnostic to whether the embeddings originate from a generic pretrained backbone or a domain-adapted self-supervised one. 
 
\subsection{Mean Shift-based Feature Refinement}
 
In this section, we introduce the iterative manifold-shifting procedure which refines feature representations by shifting latent feature vectors to regions of high empirical likelihood or density. This density step fundamentally makes our method unique since
conventional approaches directly apply density estimation in the original feature space without refining the latent space.
 
\noindent\textbf{Notation.}
Let $\mathcal{X}_{\mathrm{tr}} = \{\mathbf{x}_1,\dots,\mathbf{x}_N\} \subset \mathbb{R}^d$ denote the standardized (zero mean, unit variance) latent embeddings of the \emph{normal} training samples, and let $\mathcal{X}_{\mathrm{te}} = \{\mathbf{x}_1,\dots,\mathbf{x}_M\} \subset \mathbb{R}^d$ denote the standardized embeddings of the test samples (standardized
using the training mean and variance). Of the four stages described in the \textit{Overview} below, stages (i) and (ii) are run
\emph{twice} on different inputs: once on $\mathcal{X}_{\mathrm{tr}}$ alone during training, and once on the combined set $\mathcal{X}_{\mathrm{all}} = \mathcal{X}_{\mathrm{tr}} \cup \mathcal{X}_{\mathrm{te}}$ at inference. We mark the outputs of the training-time call with superscript $(\cdot)^{\mathrm{tr}}$ and those of the inference-time call with $(\cdot)^{\mathrm{all}}$
(e.g.\ $\mathbf{w}^{\mathrm{tr}}$ vs.\ $\mathbf{w}^{\mathrm{all}}$) so the two are never conflated. Algorithms~\ref{alg:empirical_weights} and~\ref{alg:shift_manifold} (corresponding to stages (i) and (ii) respectively) are written once, as self-contained procedures i.e. their input point set is a local parameter, written simply as $\mathcal{X} = \{\mathbf{x}_i\}_{i=1}^{n}$ inside the algorithm box (much like the argument of a function), and is bound to $\mathcal{X}_{\mathrm{tr}}$ or $\mathcal{X}_{\mathrm{all}}$ depending on which stage calls it.
 
\noindent\textbf{Overview.}
The proposed framework consists of four stages, the first three of which are performed once, at training time, exclusively on $\mathcal{X}_{\mathrm{tr}}$; the fourth is performed at
inference time on the combined population:
\begin{enumerate}\label{enum:msde_pipeline}
    \item[(i)] \emph{Empirical density weight estimation} on $\mathcal{X}_{\mathrm{tr}}$,
        producing weights $\mathbf{w}^{\mathrm{tr}}$.
    \item[(ii)] \emph{Iterative weighted manifold shifting} of $\mathcal{X}_{\mathrm{tr}}$
        using $\mathbf{w}^{\mathrm{tr}}$, producing the shifted training embeddings
        $\mathcal{X}_{\mathrm{tr}}^{*}$.
    \item[(iii)] \emph{Gaussian density estimation (GDE) fitting} on $\mathcal{X}_{\mathrm{tr}}^{*}$,
        producing a set of frozen scoring parameters.
    \item[(iv)] \emph{Transductive inference and anomaly scoring}: weights and shifted
        positions are \emph{recomputed from scratch} over $\mathcal{X}_{\mathrm{tr}} \cup
        \mathcal{X}_{\mathrm{te}}$, and the frozen GDE from stage (iii) is used to score
        \emph{only} the test embeddings.
\end{enumerate}
Stages (i) and (ii) call the same two procedures, $\textsc{EmpWeights}$ (Algorithm~\ref{alg:empirical_weights}) and $\textsc{ShiftManifold}$ (Algorithm~\ref{alg:shift_manifold}), that stage (iv) calls again later on the larger, combined population. This sharing is intentional: it guarantees that the inference-time shift uses exactly the same density-estimation and shifting logic as the training-time shift, just applied to a different point set.
 
\medskip
\noindent\textbf{(i) Empirical Density Weight Estimation.}
Each training sample $\mathbf{x}_i \in \mathcal{X}_{\mathrm{tr}}$ is assigned a scalar empirical density weight $w_i$ reflecting the local population density of its neighborhood, with samples in dense regions receiving larger weights than those in sparse or boundary
regions. We derive these weights in a data-driven manner by first constructing a fuzzy simplicial set over the $k_{\text{umap}}$-nearest-neighbor graph of $\mathcal{X}_{\mathrm{tr}}$, yielding a symmetric affinity matrix $\mathbf{G}$ whose entries encode topological proximity along the data manifold~\cite{mcinnes2018umap}. For computational tractability in high-dimensional settings, this graph is constructed in a block-wise fashion over contiguous batches of samples. A KD-tree is then built over the rows of $\mathbf{G}$, and for each sample $i$ we define the neighborhood count at radius $\epsilon$ as $c_i(\epsilon) = |\{j \neq i \mid \|\mathbf{g}_i - \mathbf{g}_j\|_2 < \epsilon\}|$.
 
Rather than fixing $\epsilon$ globally, we select it adaptively via binary search: we seek the smallest $\epsilon^*$ such that at least a fraction $\rho$ of samples exceed a density threshold $\tau$, i.e.,
\begin{equation}
    \epsilon^* = \min\left\{\epsilon \;\middle|\;
    \frac{1}{N}\sum_{i=1}^{N}\mathbf{1}\!\left[c_i(\epsilon) > \tau\right] \geq \rho\right\}.
    \label{eq:eps_search}
\end{equation}

To stabilize estimation across heterogeneous regions of the feature space, we evaluate the density count at $M_{\text{scales}} = 4$ progressively shrinking radii obtained by uniformly decrementing from $\epsilon^*$, with $\epsilon_{\min}$ fixed at $10^{-6}$. The final weight is defined as the average density count across all scales:
\begin{equation}
    w_i = \frac{1}{M_{\text{scales}}}\sum_{m=1}^{M_{\text{scales}}} c_i(\epsilon_m), \qquad
    \epsilon_m = \epsilon^* - (m-1)\Delta\epsilon, \qquad
    \Delta\epsilon = \frac{\epsilon^* - \epsilon_{\min}}{M_{\text{scales}}}.
    \label{eq:weight}
\end{equation}
Unlike kernel density estimation (KDE), which can be computationally expensive, requires careful global bandwidth selection, and degrades in high-dimensional spaces due to the curse of dimensionality, our approach operates on the low-dimensional UMAP affinity manifold and
adaptively calibrates the neighborhood radius from the data itself, yielding density estimates that are robust to the varying local geometries typical of high-dimensional anomaly detection. This is exactly the procedure $\textsc{EmpWeights}$ given in
Algorithm~\ref{alg:empirical_weights}; here it is invoked as $\mathbf{w}^{\mathrm{tr}} = \textsc{EmpWeights}(\mathcal{X}_{\mathrm{tr}}, \tau)$.

\begin{algorithm}[t]
\caption{EmpWeights: Empirical Density Weight Estimation via UMAP Graph}
\label{alg:empirical_weights}
\begin{algorithmic}[1]
\REQUIRE Point set $\mathcal{X} = \{\mathbf{x}_i\}_{i=1}^{n}$ \hfill$\triangleright$ \emph{local}
         parameter: $\mathcal{X}_{\mathrm{tr}}$ or $\mathcal{X}_{\mathrm{all}}$ at the call site;
         neighbors $k_{\text{umap}}$; count threshold $\tau$; satisfiability proportion $\rho$;
         number of scales $M_{\text{scales}}{=}4$
\ENSURE Weights $\mathbf{w} = \{w_i\}_{i=1}^{n}$
\STATE Build the $k_{\text{umap}}$-nearest-neighbor graph over $\mathcal{X}$ and the fuzzy
       simplicial set $\mathbf{G} \in \mathbb{R}^{n \times n}$~\cite{mcinnes2018umap}
       \hfill$\triangleright$ in batches of fixed size if $n$ is large
\STATE Build a KD-tree~\cite{bentley1975kdtree} over the rows $\{\mathbf{g}_i\}_{i=1}^n$ of $\mathbf{G}$
\STATE $d_{\max} \leftarrow \max_{i,j}\|\mathbf{g}_i - \mathbf{g}_j\|$,\quad
       $d_{\min} \leftarrow \min_{i \neq j}\|\mathbf{g}_i - \mathbf{g}_j\|$
\STATE $c_i(\epsilon) \leftarrow |\{j \neq i : \|\mathbf{g}_i - \mathbf{g}_j\| < \epsilon\}|$
       \hfill$\triangleright$ KD-tree range query
\STATE $\epsilon^{*} \leftarrow \textsc{BinarySearch}\Big(d_{\min}, d_{\max},\;
       \text{condition: } \tfrac{1}{n}\textstyle\sum_i \mathbf{1}[c_i(\epsilon) > \tau] \ge \rho\Big)$
\IF{search fails to satisfy the condition}
  \STATE retry with relaxed threshold $\tau/2$ and proportion $\rho/2$; if still infeasible, $\epsilon^{*} \leftarrow d_{\max}$
\ENDIF
\STATE $\Delta\epsilon \leftarrow (\epsilon^{*} - 10^{-6}) / M_{\text{scales}}$
\FOR{$m = 1$ \TO $M_{\text{scales}}$}
    \STATE $\epsilon_m \leftarrow \epsilon^{*} - (m-1)\Delta\epsilon$
    \STATE $w_i^{(m)} \leftarrow c_i(\epsilon_m) \quad \forall i = 1,\dots,n$
\ENDFOR
\STATE $\mathbf{w} \leftarrow \dfrac{1}{M_{\text{scales}}}\sum_{m=1}^{M_{\text{scales}}} \mathbf{w}^{(m)}$
\RETURN $\mathbf{w}$
\end{algorithmic}
\end{algorithm}
 
\medskip
\noindent\textbf{(ii) Iterative Weighted Manifold Shifting.}
With weights $\mathbf{w}^{\mathrm{tr}}$ established on $\mathcal{X}_{\mathrm{tr}}$, we
proceed to iterative manifold refinement of the training embeddings. At each iteration $t$,
a $k$-nearest-neighbor graph is constructed over the current sample positions to define the
local interaction structure governing feature updates. Exact $k$-NN graph construction
incurs $\mathcal{O}(N^2)$ complexity, which is prohibitive at scale. We therefore employ
NNDescent~\cite{ono2023rnn_descent}, an approximate nearest-neighbor method that exploits a
locality heuristic to build high-quality $k$-NN graphs with complexity approaching
$\mathcal{O}(N \log N)$ in practice. Denoting by $\mathcal{N}_i^{(t)}$ the set of $k$
approximate nearest neighbors of $\mathbf{x}_i$ at iteration $t$, each sample is moved
toward the weighted centroid of its local neighborhood:
\begin{equation}
    \tilde{\mathbf{x}}_i^{(t)} =
    \frac{\sum_{j \in \mathcal{N}_i^{(t)}} w_j \mathbf{x}_j^{(t)}}
         {\sum_{j \in \mathcal{N}_i^{(t)}} w_j},
    \qquad
    \mathbf{x}_i^{(t+1)} = \mathbf{x}_i^{(t)} +
    \eta \cdot \left(\tilde{\mathbf{x}}_i^{(t)} - \mathbf{x}_i^{(t)}\right),
    \label{eq:shift}
\end{equation}
where $\eta$ is a learning rate controlling the step size and the denominator in the first
expression is floored at $10^{-6}$ to avoid division by zero. This update progressively
consolidates densely-supported samples into compact manifold clusters while displacing
sparsely-supported (anomalous) samples toward more isolated positions. The process repeats
for a maximum of $T$ iterations or until the mean displacement across all training samples
falls below a threshold $\delta$:
\begin{equation}
    \frac{1}{N} \sum_{i=1}^{N}
    \|\mathbf{x}_i^{(t+1)} - \mathbf{x}_i^{(t)}\|_2 < \delta.
    \label{eq:convergence}
\end{equation}
This is exactly the procedure $\textsc{ShiftManifold}$ given in full in
Algorithm~\ref{alg:shift_manifold}; here it is invoked as $\mathcal{X}_{\mathrm{tr}}^{*} =
\textsc{ShiftManifold}(\mathcal{X}_{\mathrm{tr}}, \mathbf{w}^{\mathrm{tr}}, k, \eta, T,
\delta)$. The \emph{same} algorithm is invoked again, unmodified, in stage (iv) below on the
joint train+test population.

\begin{algorithm}[t]
\caption{ShiftManifold: Iterative Weighted Manifold Shift}
\label{alg:shift_manifold}
\begin{algorithmic}[1]
\REQUIRE Point set $\mathcal{X} = \{\mathbf{x}_i\}_{i=1}^{n}$ \hfill$\triangleright$ \emph{local}
         parameter: $\mathcal{X}_{\mathrm{tr}}$ or $\mathcal{X}_{\mathrm{all}}$ at the call site;
         weights $\mathbf{w} = \{w_i\}_{i=1}^{n}$; neighbors $k$; learning rate $\eta$; max
         iterations $T$; convergence tolerance $\delta$
\ENSURE Shifted point set $\mathcal{X}^{*} = \{\mathbf{x}_i^{*}\}_{i=1}^{n}$
\STATE $\mathcal{X}^{(0)} \leftarrow \mathcal{X}$
\FOR{$t = 1$ \TO $T$}
  \STATE $\mathcal{N}_i^{(t)} \leftarrow$ $k$ approximate nearest neighbors of $\mathbf{x}_i^{(t-1)}$
         in $\mathcal{X}^{(t-1)}$, via \textsc{NNDescent}~\cite{ono2023rnn_descent}, $\forall i$
  \FOR{each $i = 1,\dots,n$ \textbf{(in parallel)}}
    \STATE $\textit{denom} \leftarrow \max\!\Big(\sum_{j \in \mathcal{N}_i^{(t)}} w_j,\; 10^{-6}\Big)$
    \STATE $\tilde{\mathbf{x}}_i \leftarrow \Big(\sum_{j \in \mathcal{N}_i^{(t)}} w_j\, \mathbf{x}_j^{(t-1)}\Big)
           \big/ \textit{denom}$
    \STATE $\mathbf{x}_i^{(t)} \leftarrow \mathbf{x}_i^{(t-1)} + \eta\,\big(\tilde{\mathbf{x}}_i - \mathbf{x}_i^{(t-1)}\big)$
    \STATE $c_i \leftarrow \|\mathbf{x}_i^{(t)} - \mathbf{x}_i^{(t-1)}\|_2$
  \ENDFOR
  \IF{$\frac{1}{n}\sum_{i=1}^{n} c_i < \delta$}
    \STATE \textbf{break}
  \ENDIF
\ENDFOR
\RETURN $\mathcal{X}^{*} \leftarrow \mathcal{X}^{(t)}$
\end{algorithmic}
\end{algorithm}
 
\medskip
\noindent\textbf{(iii) Gaussian Density Estimation Fitting.}
Once the training embeddings have converged to $\mathcal{X}_{\mathrm{tr}}^{*} = \{\mathbf{x}_i^{*}\}_{i=1}^{N}$, we fit a Gaussian density estimator (GDE) on this shifted training population. To mitigate the curse of dimensionality and stabilize the covariance estimate, we first project onto a $d' = \min(256, d, N{-}1)$-dimensional PCA subspace fit on $\mathcal{X}_{\mathrm{tr}}^{*}$, yielding projection matrix $\mathbf{P}$ and mean $\boldsymbol{\mu}_{\mathrm{pca}}$. Writing $\mathbf{z}_i = \mathbf{P}^\top(\mathbf{x}_i^{*} - \boldsymbol{\mu}_{\mathrm{pca}})$ for the projected training points, we estimate
\begin{equation}
    \boldsymbol{\mu}_z = \frac{1}{N}\sum_{i=1}^{N}\mathbf{z}_i, \qquad
    \boldsymbol{\Sigma}_z = \frac{1}{N-1}\sum_{i=1}^{N}
    (\mathbf{z}_i - \boldsymbol{\mu}_z)(\mathbf{z}_i - \boldsymbol{\mu}_z)^\top +
    \lambda \mathbf{I},
    \label{eq:gde_fit}
\end{equation}
where $\lambda = 10^{-4}$ regularizes the covariance to keep it invertible in
high dimensions. The quadruple $\theta_{\mathrm{GDE}} = (\mathbf{P}, \boldsymbol{\mu}_{\mathrm{pca}},
\boldsymbol{\mu}_z, \boldsymbol{\Sigma}_z^{-1})$ is computed once, from the shifted training
embeddings only, and is then \emph{frozen}: it is reused unchanged at inference time and is
never refit on test data.
 
\medskip
\noindent\textbf{(iv) Transductive Inference and Anomaly Scoring.}
To ensure geometric consistency between training and test representations, manifold shifting is applied jointly at inference time to $\mathcal{X}_{\mathrm{all}} = \mathcal{X}_{\mathrm{tr}} \cup \mathcal{X}_{\mathrm{te}}$, where $\mathcal{X}_{\mathrm{tr}}$
here denotes the \emph{original, unshifted} training embeddings (not $\mathcal{X}_{\mathrm{tr}}^{*}$ from stage (ii)). This is a transductive refinement step: test samples participate in neighborhood construction so that their shifted positions are evaluated in the same manifold context as training samples, preventing the representation mismatch that would arise from shifting the two splits independently. Concretely,
\begin{equation}
    \mathbf{w}^{\mathrm{all}} = \textsc{EmpWeights}(\mathcal{X}_{\mathrm{all}}, \tau), \qquad
    \mathcal{X}_{\mathrm{all}}^{*} = \textsc{ShiftManifold}(\mathcal{X}_{\mathrm{all}},
    \mathbf{w}^{\mathrm{all}}, k, \eta, T, \delta),
    \label{eq:joint_shift}
\end{equation}
i.e.\ both the empirical weights \emph{and} the shifted positions are recomputed from
scratch over the combined population, and are generally different from $\mathbf{w}^{\mathrm{tr}}$
and $\mathcal{X}_{\mathrm{tr}}^{*}$ obtained in stages (i)-(ii). Let
$\mathcal{X}_{\mathrm{te}}^{*} = \{\mathbf{x}_j^{*}\}_{j=1}^{M}$ denote the slice of
$\mathcal{X}_{\mathrm{all}}^{*}$ corresponding to the test samples. Crucially, scoring is
performed \emph{only} on this test slice, using the GDE parameters $\theta_{\mathrm{GDE}}$
frozen in stage (iii): for each test point we project $\mathbf{z}_j = \mathbf{P}^\top(\mathbf{x}_j^{*}
- \boldsymbol{\mu}_{\mathrm{pca}})$ and compute the Mahalanobis distance
\begin{equation}
    s_j = \sqrt{(\mathbf{z}_j - \boldsymbol{\mu}_z)^\top \boldsymbol{\Sigma}_z^{-1}
    (\mathbf{z}_j - \boldsymbol{\mu}_z)}, \qquad j = 1,\dots,M.
    \label{eq:maha}
\end{equation}
Training points are never re-scored at inference; their sole role in stage (iv) is to stabilize the local manifold geometry around the test points. The raw distances $\{s_j\}_{j=1}^{M}$ are finally standardized using their own sample mean $\hat{\mu}_s$ and standard deviation $\hat{\sigma}_s$ and passed through a sigmoid to obtain calibrated, ranking-preserving scores
in $(0,1)$:
\begin{equation}
    s_j \leftarrow \sigma\!\left(\frac{s_j - \hat{\mu}_s}{\hat{\sigma}_s}\right), \qquad j = 1,\dots,M.
    \label{eq:calibration}
\end{equation}
Critically, this transductive design does not constitute information leakage: no test labels are used at any stage, $\theta_{\mathrm{GDE}}$ is fitted exclusively on the shifted training embeddings and remains frozen, and the calibration in Eq.~\eqref{eq:calibration} uses only the (label-free) test distances themselves. 
 
\begin{algorithm}[t]
\caption{Manifold Shift Density Estimator (end-to-end)}
\label{alg:msde}
\small
\begin{algorithmic}[1]
\REQUIRE $\mathcal{X}_{\mathrm{tr}} = \{\mathbf{x}_i\}_{i=1}^{N}$ (standardized, normal-only
         training embeddings); $\mathcal{X}_{\mathrm{te}} = \{\mathbf{x}_j\}_{j=1}^{M}$
         (standardized test embeddings); shift params $k, \tau, \eta, T, \delta$
         ($k_{\mathrm{umap}}{=}15$ fixed); PCA cap $d_{\max}{=}256$; covariance
         regularizer $\lambda{=}10^{-4}$
\ENSURE Anomaly scores $\mathbf{s} \in (0,1)^{M}$ for $\mathcal{X}_{\mathrm{te}}$ \emph{only}
 
\STATE \textbf{Stage (i)+(ii) Train-only manifold shift}
\STATE $\mathbf{w}^{\mathrm{tr}} \leftarrow \textsc{EmpWeights}(\mathcal{X}_{\mathrm{tr}}, \tau)$
       \hfill$\triangleright$ Alg.~\ref{alg:empirical_weights}
\STATE $\mathcal{X}_{\mathrm{tr}}^{*} \leftarrow \textsc{ShiftManifold}(\mathcal{X}_{\mathrm{tr}},
       \mathbf{w}^{\mathrm{tr}}, k, \eta, T, \delta)$
       \hfill$\triangleright$ Alg.~\ref{alg:shift_manifold}
 
\STATE \textbf{Stage (iii) Fit GDE on shifted training embeddings}
\STATE $d' \leftarrow \min(d_{\max}, d, N{-}1)$;\quad
       fit $\mathrm{PCA}(d')$ on $\mathcal{X}_{\mathrm{tr}}^{*}$
       $\Rightarrow \mathbf{P}, \boldsymbol{\mu}_{\mathrm{pca}}$
\STATE $\mathbf{z}_i \leftarrow \mathbf{P}^\top(\mathbf{x}_i^{*} - \boldsymbol{\mu}_{\mathrm{pca}})
       \ \forall\, \mathbf{x}_i^{*} \in \mathcal{X}_{\mathrm{tr}}^{*}$;\quad
       $\boldsymbol{\mu}_z \leftarrow \tfrac{1}{N}\sum_i \mathbf{z}_i$
\STATE $\boldsymbol{\Sigma}_z \leftarrow \tfrac{1}{N-1}\sum_i (\mathbf{z}_i - \boldsymbol{\mu}_z)
       (\mathbf{z}_i - \boldsymbol{\mu}_z)^\top + \lambda \mathbf{I}$;\quad
       $\boldsymbol{\Sigma}_z^{-1} \leftarrow \mathrm{inv}(\boldsymbol{\Sigma}_z)$
\STATE \emph{Freeze} $\theta_{\mathrm{GDE}} \leftarrow (\mathbf{P}, \boldsymbol{\mu}_{\mathrm{pca}},
       \boldsymbol{\mu}_z, \boldsymbol{\Sigma}_z^{-1})$ \hfill$\triangleright$ reused unchanged below
 
\STATE \textbf{Stage (iv) Transductive joint shift, scoring \emph{test points only}}
\STATE $\mathcal{X}_{\mathrm{all}} \leftarrow \mathcal{X}_{\mathrm{tr}} \cup \mathcal{X}_{\mathrm{te}}$
       \hfill$\triangleright$ \emph{original, unshifted} train embeddings, rejoined with test
\STATE $\mathbf{w}^{\mathrm{all}} \leftarrow \textsc{EmpWeights}(\mathcal{X}_{\mathrm{all}}, \tau)$
       \hfill$\triangleright$ recomputed from scratch; $\mathbf{w}^{\mathrm{all}} \neq \mathbf{w}^{\mathrm{tr}}$
\STATE $\mathcal{X}_{\mathrm{all}}^{*} \leftarrow \textsc{ShiftManifold}(\mathcal{X}_{\mathrm{all}},
       \mathbf{w}^{\mathrm{all}}, k, \eta, T, \delta)$
\STATE $\mathcal{X}_{\mathrm{te}}^{*} \leftarrow \mathcal{X}_{\mathrm{all}}^{*}[N : N{+}M]$
       \hfill$\triangleright$ 0-based slice; \emph{only} the test positions are kept
\FOR{$j = 1$ \TO $M$}
  \STATE $\mathbf{z}_j \leftarrow \mathbf{P}^\top(\mathbf{x}_j^{*} - \boldsymbol{\mu}_{\mathrm{pca}})$
         \hfill$\triangleright$ test indices only; using frozen $\theta_{\mathrm{GDE}}$ from Stage (iii)
  \STATE $s_j \leftarrow \sqrt{(\mathbf{z}_j - \boldsymbol{\mu}_z)^\top \boldsymbol{\Sigma}_z^{-1}
         (\mathbf{z}_j - \boldsymbol{\mu}_z)}$
\ENDFOR
\STATE $\hat{\mu}_s, \hat{\sigma}_s \leftarrow$ mean, std.\ of $\{s_j\}_{j=1}^{M}$
       \hfill$\triangleright$ test scores only; ranking preserved
\STATE $s_j \leftarrow \sigma\!\left(\dfrac{s_j - \hat{\mu}_s}{\hat{\sigma}_s}\right) \quad \forall j = 1,\dots,M$
\RETURN $\mathbf{s} = \{s_j\}_{j=1}^{M}$
\end{algorithmic}
\end{algorithm}

\section{Experimental details}

\subsection{Datasets}
 To ensure fair comparison and reproducibility, we strictly follow the dataset construction, preprocessing, and evaluation protocol defined in the MedIAnomaly benchmark~\cite{CAI2025103500}. We evaluate our proposed method on seven publicly available medical imaging datasets spanning five imaging modalities, including chest X-rays, brain MRI, retinal fundus images, dermatoscopic images, and histopathology images, as summarized in Table \ref{tab:dataset_summary}.

The \textit{RSNA Pneumonia Detection Dataset} (RSNA)~\cite{stein2018rsna} and \textit{VinDr-CXR Dataset} (Vin)~\cite{nguyen2022vindrcxr} consist of chest X-ray images, where normal cases are used for training and images with pneumonia or other pathological findings are treated as anomalies during testing. The \textit{Brain Tumor Dataset} (Brain)~\cite{nickparvar2021brain} and \textit{BraTS2021 Dataset} (BraTS)~\cite{baid2021brats} involve brain MRI data; in both cases, slices without tumors are considered normal, while tumor-containing slices are treated as anomalous. 

The \textit{LAG Dataset} (LAG)~\cite{li2019lag} focuses on retinal fundus images for glaucoma detection, where normal images are used for training and glaucomatous images are considered anomalies. The \textit{ISIC 2018 Dataset} (ISIC)~\cite{codella2019isic} contains dermatoscopic images, where nevus (NV) is defined as the normal class and all other lesion categories are treated as anomalous. The \textit{Camelyon16 Dataset} (C16)~\cite{litjens2018camelyon16} comprises histopathology whole-slide images, from which patches of size $256 \times 256$ at 40$\times$ magnification are extracted; patches containing tumor regions are considered anomalies.

Across all datasets, we adopt an OCC-based anomaly detection setting as defined in MedIAnomaly \cite{CAI2025103500}, where only normal samples are used during training, and evaluation is performed on a mixture of normal and abnormal samples using the same splits and preprocessing pipeline provided in the benchmark.

\begin{table}[t]
\centering
\small
\setlength{\tabcolsep}{4pt}
\renewcommand{\arraystretch}{1.2}
\begin{tabular}{|p{2cm}|p{2.7cm}|p{2.7cm}|p{2cm}|p{1.8cm}|p{2cm}|}
\hline
\textbf{Dataset} & \textbf{Modality / Task} & \textbf{Feature Learning Backbone} & \textbf{Train (Normal)} & \textbf{Test (Normal)} & \textbf{Test (Abnormal)} \\
\hline

RSNA Pneumonia & Chest X-ray / Pneumonia Detection & AnatPaste & 3,851 & 1,000 & 1,000 \\
\hline

VinDr-CXR & Chest X-ray / Multi-abnormality Detection & AnatPaste & 4,000 & 1,000 & 1,000 \\
\hline

Brain Tumor & MRI / Tumor Detection & ResNet18 & 1,000 & 600 & 600 \\
\hline

LAG & Fundus / Glaucoma Detection & ResNet18 & 1,500 & 811 & 811 \\
\hline

ISIC 2018 & Dermoscopy / Skin Lesion Analysis & AnatPaste & 6,705 & 909 & 603 \\
\hline

Camelyon16 & Histopathology / Metastasis Detection & ResNet18 & 5,088 & 1,120 & 1,113 \\
\hline

BraTS2021 & MRI (FLAIR) / Tumor Segmentation & ResNet18 & 4,211 & 828 & 1,948 \\
\hline

\end{tabular}
\caption{Summary of datasets used for anomaly detection. Only normal samples are used for training, while test sets contain both normal and abnormal samples.}
\label{tab:dataset_summary}
\end{table}

For all datasets, we strictly follow the data splits and preprocessing pipeline provided in the MedIAnomaly repository. The training set contains only normal samples, while the test set includes both normal and anomalous samples. All images are resized to a fixed resolution and normalized prior to feature extraction. Grayscale images are converted to three-channel format (RGB) to ensure compatibility with pre-trained convolutional neural network backbones.

This diverse collection of datasets enables a comprehensive evaluation of the proposed method across different modalities, anatomical regions, and anomaly characteristics.

\subsection{Selection of Baseline Models}

To ensure a focused comparison, we select a subset of representative methods based on their superior performance in the original benchmark (see Table 10 in~\cite{CAI2025103500}), prioritizing models that achieve the highest number of top-ranking results (best, second-best, and third-best) across datasets, as well as those with publicly available implementations. Based on these selection criteria, we select the models AE-PL, AE-U, AnatPaste, CutPaste-3way, CutPaste, ResNet152, and ResNet18. It is worth noting that certain methods with good performance reported in the original benchmark, such as MSC~\cite{reiss2023msc} and PANDA~\cite{reiss2021panda}, could not be included in our benchmarking studies due to the absence of their implementations in the official repository.
We note that Tables~\ref{tab:msde_AUC_comparison} and~\ref{tab:msde_AP_comparison} additionally report, for each dataset, the single best-performing method and score from the full original MedIAnomaly benchmark~\cite{CAI2025103500}, which may include methods such as MSC~\cite{reiss2023msc}, DAE~\cite{kascenas2022dae}, AutoDDPM~\cite{bercea2023autoddpm}, FAE-SSIM~\cite{ristea2022self}, and f-AnoGAN~\cite{schlegl2019fanogan} that we did not reproduce ourselves; these figures are taken directly from the published benchmark rather than from our own runs, and are included for completeness alongside our own reproduced baselines in Table~\ref{tab:benchmark_results}.

\subsection{Implementation Details}\label{sec:implementation_details}

The proposed framework is implemented by extending the MedIAnomaly pipeline and integrating the novel Mean Shift-based Feature Refinement module within its feature processing and scoring stages. This section summarizes the practical configuration used for all experiments.

\paragraph{Feature Embeddings.}
We utilize pre-extracted feature representations provided by the MedIAnomaly setup. For the datasets RSNA, VinDr-CXR, and ISIC2018, embeddings are obtained using the AnatPaste self-supervised model, while for all other datasets, a ResNet18 backbone pretrained on ImageNet is employed. In both cases, each image is represented as a 512-dimensional feature vector. Feature extraction is performed offline, and the same embeddings are reused across all experiments to ensure consistency and computational efficiency. More details on the default parameters can be found in the pseudocodes Algorithm \ref{alg:msde}.

\paragraph{Evaluation Protocol.}
Performance is evaluated using AUC-ROC and Average Precision (AP), following standard anomaly detection benchmarks~\cite{CAI2025103500}. All experiments are conducted under a one-class training setting using only normal samples. Hyperparameters of our model, including the number of neighbors, density threshold, learning rate, and convergence criteria, are reported in Table~\ref{tab:msde_hyperparameters}.

All experiments are implemented in Python using PyTorch and standard scientific libraries, and executed on a GPU-enabled computing environment.

\begin{table*}[t]
\centering
\caption{Fixed hyperparameters used for the proposed model across all datasets.}
\label{tab:msde_hyperparameters}
\begin{tabular}{lcp{5.8cm}}
\toprule
\textbf{Hyperparameter} & \textbf{Value} & \textbf{Description} \\
\midrule
\texttt{Number of Neighbors} ($k$) & 50 & Number of nearest neighbors used to construct the local manifold graph \\
\texttt{Neighborhood Sample Threshold} ($\tau$) & 70 & Minimum number of neighbors required inside the estimated radius during weight computation. Controls density sensitivity \\
\texttt{Learning Rate} ($\eta$) & 0.33 & Step size controlling the magnitude of each mean shift update \\
\texttt{Max Shift Iterations} ($T$) & 8 & Maximum number of iterations for the manifold refinement process \\
\texttt{Shift Convergence Threshold} ($\delta$) & 0.01 & Stopping criterion based on average shift magnitude between iterations \\
\bottomrule
\end{tabular}
\end{table*}

\section{Results}
\label{sec:results}

In this section, we present the results of our Benchmarking experiment. We show that our pipeline consistently outperforms state-of-the-art models across both AUC and AP.


\paragraph{Comparative Analysis Against State-of-the-Art Baselines.}

As detailed in Table~\ref{tab:msde_results}-Table~\ref{tab:benchmark_results}, even without dataset-specific tuning, the proposed pipeline delivers exceptional baseline performance. For the Brain Tumor dataset, our model achieves a score of 0.981 for both AUC and AP simultaneously. In the historical context of medical OCC, models often suffer from a strict precision-recall trade-off. A model optimized for high AUC (general ranking) typically sacrifices AP (precision against false positives). The results demonstrate that our model performs competitively across both metrics when compared with existing state-of-the-art approaches.

Table~\ref{tab:msde_AUC_comparison} and Table~\ref{tab:msde_AP_comparison} show that our model achieves state-of-the-art AUC performance on 4 out of the 7 evaluated datasets (RSNA, VinDr-CXR, Brain Tumor, Camelyon16), and achieves the highest AP on 5 out of the 7 datasets. We would like to point out that such cross-modality consistency is rarely observed among the models compared in the recent work of Cai \textit{et al.} \cite{CAI2025103500}. 

We observe from the work of Cai \textit{et al.} that reconstruction models like AE-PL dominate specific tasks like retinal imagery (LAG), while generative diffusion models like AutoDDPM excel at highly textured histopathology (Camelyon16). Our method however shows superior performance on both. 

The comparison on the Camelyon16 gigapixel dataset is particularly noteworthy. AutoDDPM  leverages computationally intensive diffusion resampling processes to achieve an AUC of 0.807 and an AP of 0.767. Our algorithm, operating purely on extracted ResNet18 features via relatively lightweight geometric shifting, surpasses this diffusion baseline, achieving an AUC of 0.812 and an AP of 0.820.

Furthermore, it is also interesting to observe that we outperform the sophisticated feature-reconstruction autoencoders. In the VinDr-CXR multi-pathology benchmark, the previous best metric was held by the Uncertainty Autoencoder (AE-U) with an AUC of 0.769. Our method improves upon this with an AUC of 0.819.

On the ISIC 2018 dermoscopy dataset we observe that our model underperforms compared to the fully optimized AnatPaste model (0.705 vs.\ 0.807 AUC). This behavior is consistent with the underlying assumptions of the feature extractor. AnatPaste is designed to exploit structured anatomical regions (like distinct lung segmentations), whereas dermoscopic images exhibit relatively homogeneous textures without well-defined internal structures. Consequently, the extracted features are less discriminative, and the mean shift based method inherits this limitation because its performance depends on the quality and informativeness of the underlying feature representations.

\begin{table}[t]
\centering
\caption{Performance of our model with fixed hyperparameters across all datasets.}
\label{tab:msde_results}
\begin{tabular}{lccc}
\hline
\textbf{Dataset} & \textbf{Backbone} & \textbf{AUC} & \textbf{AP} \\
\hline
RSNA  & AnatPaste & 0.918 & 0.906 \\
Vin   & AnatPaste & 0.819 & 0.797 \\
ISIC  & AnatPaste & 0.705 & 0.638 \\
Brain & ResNet18  & 0.981 & 0.981 \\
LAG   & ResNet18  & 0.810 & 0.831 \\
BraTS   & ResNet18  & 0.736 & 0.867 \\
C16   & ResNet18  & 0.812 & 0.820 \\

\hline
\end{tabular}
\end{table}

\begin{table}[t]
\centering
\caption{Comparison of AUC of our model with the best existing methods on each dataset from the MedIAnomaly paper\cite{CAI2025103500}(Page~10). We have used a fixed set of hyperparameters across all datasets. Best results are highlighted in bold.}
\label{tab:msde_AUC_comparison}
\begin{tabular}{lcccc}
\hline
\textbf{Dataset} & \textbf{Best Existing Model} & \textbf{Best Existing} & \textbf{Ours} \\
\hline
RSNA  & FAE-SSIM~\cite{ristea2022self} & 0.911 & \textbf{0.918}  \\
Vin   & AE-U~\cite{mao2020aeu}  & 0.769 & \textbf{0.819}  \\
ISIC  & AnatPaste~\cite{sato2023anatomy}  & \textbf{0.807} & 0.705  \\
Brain & MSC~\cite{reiss2023msc}  & 0.973 & \textbf{0.981}  \\
LAG   & AE-PL~\cite{shvetsova2021aepl}  & \textbf{0.856} & 0.810  \\
BraTS & DAE~\cite{kascenas2022dae} & \textbf{0.859} & 0.736  \\
C16   & AutoDDPM~\cite{bercea2023autoddpm}  & 0.807 & \textbf{0.812}  \\
\hline
\end{tabular}
\end{table}

\begin{table}[t]
\centering
\caption{Comparison of AP of our model with the best existing methods on each dataset from the MedIAnomaly paper\cite{CAI2025103500}(Page~10). We used a fixed set of hyperparameters across all datasets. Best results are highlighted in bold.}
\label{tab:msde_AP_comparison}
\begin{tabular}{lcccc}
\hline
\textbf{Dataset} & \textbf{Best Existing Model} & \textbf{Best Existing} & \textbf{Ours} \\
\hline
RSNA  & FAE-SSIM~\cite{ristea2022self} & 0.892 & \textbf{0.906}  \\
Vin   & AE-U~\cite{mao2020aeu}  & 0.754 & \textbf{0.797}  \\
ISIC  & AnatPaste~\cite{sato2023anatomy}  & \textbf{0.681} & 0.638 \\
Brain & ResNet152-IN~\cite{he2016residual}  & 0.963 & \textbf{0.981}  \\
LAG   & AE-PL~\cite{shvetsova2021aepl}  & 0.816 & \textbf{0.831} \\
BraTS & f-AnoGAN~\cite{schlegl2019fanogan}  & \textbf{0.935} & 0.867  \\
C16   & AutoDDPM~\cite{bercea2023autoddpm}  & 0.767 & \textbf{0.820}  \\
\hline
\end{tabular}
\end{table}

Table~\ref{tab:benchmark_results} presents the performance of top selected baseline methods reproduced from the MedIAnomaly benchmark~\cite{CAI2025103500}. The results are obtained by running the official implementations provided in the corresponding repository~\cite{medianomaly_repo}, ensuring consistency with the original experimental setup. In Table~\ref{tab:benchmark_results} for better interpretability, we highlight the \best{best}, \second{second-best}, and \third{third-best} results in each column using red, blue, and green colors, respectively. To summarize, the proposed method consistently achieves competitive and state-of-the-art performance across the evaluated datasets.

\begin{table*}[ht]
\centering
\caption{Image-level anomaly classification (AUC / AP \%). Results reproduced using the official MedIAnomaly implementation~\cite{medianomaly_repo}, following the experimental protocol described in~\cite{CAI2025103500}.}
\label{tab:benchmark_results}
\resizebox{\textwidth}{!}{%
\begin{tabular}{lcccccccccccccc}
\toprule
\textbf{Method} & \multicolumn{2}{c}{\textbf{RSNA}} & \multicolumn{2}{c}{\textbf{VinDr-CXR}} & \multicolumn{2}{c}{\textbf{Brain}} & \multicolumn{2}{c}{\textbf{LAG}} & \multicolumn{2}{c}{\textbf{BraTS2021}} & \multicolumn{2}{c}{\textbf{Camelyon16}} & \multicolumn{2}{c}{\textbf{ISIC2018}} \\
\cmidrule(lr){2-3} \cmidrule(lr){4-5} \cmidrule(lr){6-7} \cmidrule(lr){8-9} \cmidrule(lr){10-11} \cmidrule(lr){12-13} \cmidrule(lr){14-15}
 & \textbf{AUC} & \textbf{AP} & \textbf{AUC} & \textbf{AP} & \textbf{AUC} & \textbf{AP} & \textbf{AUC} & \textbf{AP} & \textbf{AUC} & \textbf{AP} & \textbf{AUC} & \textbf{AP} & \textbf{AUC} & \textbf{AP} \\
\midrule
Ours & \best{91.8} & \best{90.6} & \best{81.9} & \best{79.7} & \best{98.1} & \best{98.1} & \third{81.0} & \best{83.1} & 73.6 & \third{86.7} & \best{81.2} & \best{82.0} & 70.5 & \second{63.8} \\
AE-PL & \third{87.9} & \second{85.4} & \second{75.4} & \second{73.3} & \third{95.7} & 91.8 & \best{85.9} & \third{81.8} & \best{86.1} & \best{92.9} & \second{76.1} & \third{67.9} & 68.1 & 51.8 \\
AE-U & 86.4 & 84.0 & \third{73.9} & \second{73.3} & 94.0 & 88.2 & \second{84.2} & \second{82.1} & \second{85.9} & \second{89.6} & 59.3 & 54.7 & 68.4 & 57.3 \\
AnatPaste (2-stage) & \second{88.0} & \third{84.6} & 68.4 & \third{68.5} & 95.5 & 91.1 & 74.0 & 68.6 & 73.4 & 82.0 & 52.3 & 49.4 & \best{80.5} & \best{67.0} \\
CutPaste-3way (2-stage) & 77.4 & 74.3 & 57.4 & 55.5 & 89.3 & 86.0 & 67.7 & 63.3 & 52.8 & 69.8 & 47.2 & 49.8 & \third{75.1} & 61.0 \\
CutPaste (2-stage) & 69.3 & 68.8 & 55.8 & 55.0 & 93.9 & 90.8 & 70.6 & 67.6 & 52.4 & 69.6 & 41.8 & 45.8 & 71.8 & 57.5 \\
ResNet152 & 84.8 & 82.8 & 52.7 & 53.7 & \second{97.2} & \second{96.3} & 76.0 & 72.2 & 67.0 & 81.7 & \third{74.6} & \second{69.1} & \second{75.2} & \third{62.4} \\
ResNet18 & 81.1 & 79.5 & 58.6 & 57.8 & \third{95.7} & \third{93.9} & 80.8 & 77.4 & \third{75.7} & 86.1 & 73.1 & 63.8 & 67.1 & 54.1 \\
\bottomrule
\end{tabular}}
\end{table*}

\begin{figure}[t]
    \centering
    \includegraphics[width=\linewidth]{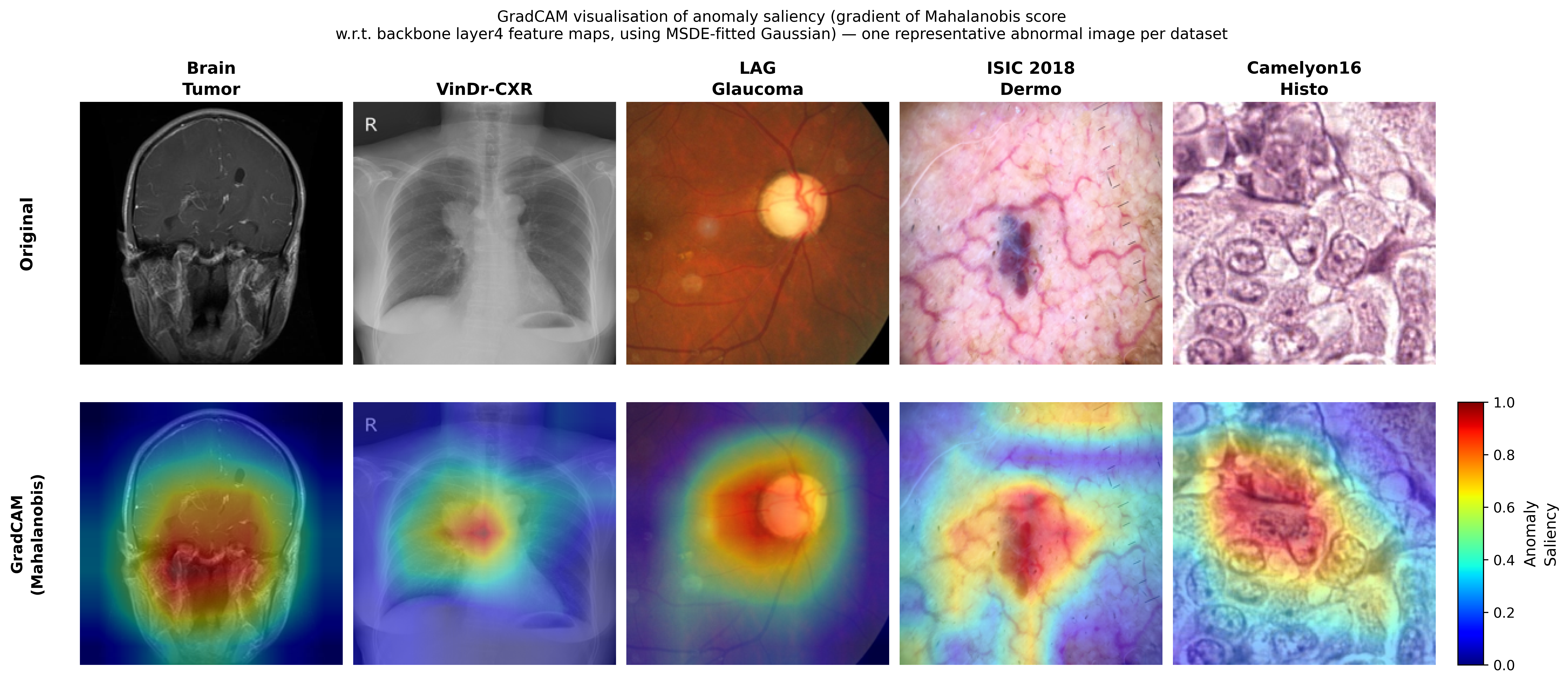}
    \caption{GradCAM visualizations of anomaly saliency across all five modalities (one per modality). Top row: representative abnormal test images. Bottom row: GradCAM heatmaps obtained by backpropagating the  Mahalanobis anomaly score (computed using the models-fitted Gaussian parameters) through the frozen backbone to its last convolutional layer (\texttt{layer4}). Warmer colors indicate higher contribution to the anomaly score. The backbone used per dataset reflects the feature extraction protocol: AnatPaste for VinDr-CXR and ISIC~2018, and ImageNet-pretrained ResNet18 for Brain Tumor, LAG, and Camelyon16. Saliency consistently localizes to clinically relevant pathological regions without any explicit localization supervision, consistent with the one-class training setting.}
    \label{fig:gradcam}
\end{figure}

\paragraph{Qualitative Analysis of Spatial Anomaly Attribution via GradCAM.}
To obtain qualitative insights on the performance of our algorithm, we investigate which spatial regions of the input image contribute most to the anomaly score under our framework. The mathematical details are provided in~\ref{gradcam}. Figure~\ref{fig:gradcam} presents GradCAM visualizations for representative abnormal test images across the five benchmark modalities. Since the proposed framework is trained solely for image-level anomaly detection and does not use any localization supervision, the GradCAM visualizations should be interpreted as qualitative indicators of regions contributing to the anomaly score rather than precise lesion segmentations. Consequently, spatial correspondence between the heatmaps and pathological regions is not guaranteed on a per-image basis.

Nevertheless, it is encouraging to observe that, despite being trained solely for image-level anomaly detection, the resulting saliency maps sometimes highlight anatomically and clinically relevant regions. While not intended as a quantitative localization assessment, these qualitative examples suggest that the anomaly score is influenced by features associated with the underlying pathology rather than by irrelevant image content. We observe that:
\begin{itemize}
    \item For the representative example from VinDr-CXR chest X-ray dataset we observe that, the heatmap highlights the lung parenchyma and mediastinal regions, which are the anatomically relevant areas for thoracic abnormalities.

    \item In the example from brain MRI dataset, which corresponds to Brain Tumor detection, saliency is concentrated on the tumor mass and surrounding tissue, demonstrating that the anomaly attribution aligns with pathological ground truth.

    \item For the example from LAG glaucoma dataset, the model focuses on the optic disc region, which is the primary site of glaucomatous damage in retinal fundus images.

    \item In the example from ISIC 2018 dermoscopy dataset, the model attends to the lesion boundary and central texture, consistent with melanocytic abnormality.

    \item In the example from Camelyon16 histopathology dataset, the model attends to regions of tissue with altered cellular architecture, consistent with metastatic tumor infiltration.
\end{itemize}
Overall, these examples provide qualitative evidence that the anomaly score is influenced by clinically relevant image content rather than arbitrary background features. Although not a substitute for a dedicated localization evaluation, the observed attribution patterns support the interpretability of the proposed framework across multiple imaging modalities.

%

\begin{figure}[t]
    \centering
    \includegraphics[width=\linewidth]{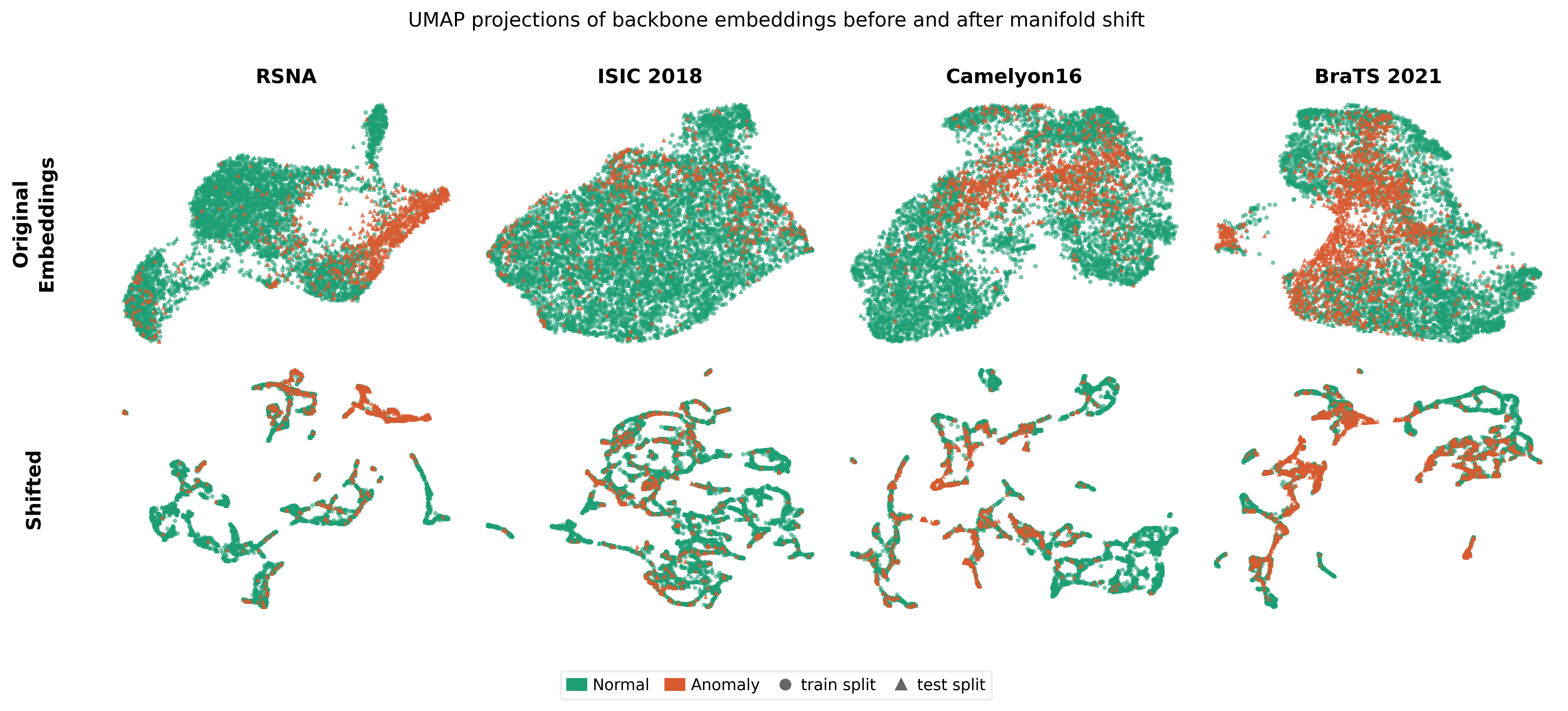}
    \caption{UMAP projections of backbone embeddings before (top row) and after (bottom row) manifold shift, across four datasets and two backbone architectures. Columns from left to right: RSNA Pneumonia and ISIC~2018 (AnatPaste backbone), Camelyon16 and BraTS~2021 (ResNet18 backbone). Green and orange markers denote normal and anomalous samples respectively; circles indicate training split and triangles indicate test split. After shifting, normal samples consistently collapse into compact topological hubs while anomalous samples are displaced to the periphery, demonstrating improved geometric separability across diverse modalities and backbones without any anomaly supervision.}
    \label{fig:umap_viz}
\end{figure}

\paragraph{Qualitative Analysis of Feature Space Manifold.}
To further investigate the effect of the proposed Mean Shift-based Feature Refinement module, we visualize the feature space before and after manifold shifting using Uniform Manifold Approximation and Projection (UMAP). The high-dimensional embeddings are projected into a two-dimensional space for qualitative inspection. Figure~\ref{fig:umap_viz} shows the resulting visualizations for four datasets spanning two backbone architectures: RSNA and ISIC~2018 using the AnatPaste backbone, and Camelyon16 and BraTS~2021 using a ResNet18 backbone.

In the original feature space (top row), normal and anomalous samples are often intermingled, and the underlying manifold structure appears relatively diffuse. After applying the proposed manifold shift (bottom row), the embeddings exhibit a noticeably different geometric organization, characterized by more structured and locally coherent manifold patterns. Across all datasets, the transformed embeddings form distinct low-dimensional structures and regions of increased local density, indicating that the shift operation substantially reshapes the feature geometry while preserving neighborhood relationships.

Although normal and anomalous samples are not perfectly separated in the UMAP projections, they more frequently occupy different regions of the transformed manifold than in the original embedding space. This behavior is consistent across datasets and backbone architectures, suggesting that the proposed refinement promotes a feature representation that is more amenable to density- and distance-based anomaly scoring. While UMAP visualizations provide only a qualitative view of the underlying high-dimensional space, the observed geometric reorganization is consistent with the intended effect of the manifold shift: concentrating samples along locally dense regions of the feature manifold and enhancing its structural organization for subsequent Mahalanobis-based anomaly detection.

\section{Discussion}

\paragraph{Implications of Latent Manifold Refinement for Anomaly Detection.}
The strong performance of our model without any retraining of the underlying backbone network indicates that pretrained latent representations often contain substantial unused discriminative information. In many anomaly detection pipelines, extracted features are treated as final inputs for downstream scoring methods. Our results suggest that these embeddings may still possess hidden geometric structure that is not fully exploited in their original form. By refining the latent manifold through density-guided shifting, the proposed framework is able to reveal and utilize this information more effectively. This finding implies that improvements in anomaly detection can arise not only from better encoders, but also from better post-processing of existing representations.

\paragraph{Broad adaptability through modular design.}
A major practical advantage of the proposed framework is its modular design, which allows it to be integrated with a wide range of existing feature extractors. Because the method operates entirely after representation learning, it does not require modifications to upstream backbone architectures or retraining procedures. This makes it compatible with standard convolutional networks, self-supervised medical encoders, and emerging foundation models. As stronger pretrained models continue to appear, the method can be directly applied as an additional refinement stage. Such flexibility increases its long-term relevance and makes it attractive for deployment in real-world medical imaging systems with diverse modalities and evolving model ecosystems.

\paragraph{Modality-agnostic deployment for rapid clinical screening.}
Because the framework operates entirely downstream of feature extraction, adapting it to a new modality requires only a corresponding pretrained backbone, with no architectural changes or modality-specific tuning. This is unlike methods that rely on anatomical priors or modality-tailored designs, e.g., diffusion models for textured histopathology, or anatomy-aware pretext tasks for chest radiographs. We confirm this in practice: the same manifold-refinement and scoring pipeline, with identical hyperparameters (Table~\ref{tab:msde_hyperparameters}), achieves state-of-the-art or competitive performance across chest X-ray, brain MRI, retinal fundus, dermatoscopy, and histopathology data. Combined with its label-free, one-class training, this makes the framework well suited as a first-line screening tool, flagging potentially abnormal cases for specialist review before modality-specific workup, without requiring per-department retraining or abnormal-case annotations.

\paragraph{Computational Efficiency and Scalability.}
The proposed framework operates on pre-extracted feature embeddings rather than raw images, avoiding expensive end-to-end retraining. The dominant computational cost arises from approximate nearest-neighbor graph construction and iterative neighborhood updates over $T$ refinement steps. Using NN-Descent~\cite{ono2023rnn_descent}, graph construction scales approximately as $\mathcal{O}(N\log N)$ in practice, while each refinement step requires $\mathcal{O}(NkD)$ operations, where $N$ is the number of samples, $D$ the feature dimension, and $k$ the neighborhood size. Since $T$ is small ($T=8$ in all experiments), the method remains computationally practical. After refinement, PCA and Gaussian scoring introduce comparatively minor overhead. Overall, the proposed framework is substantially lighter than reconstruction or diffusion-based anomaly detection methods that require repeated image-space generation or network optimization.

\paragraph{Limitation.}
Despite its strong overall performance, our framework has two notable limitations. First, it remains dependent on the quality of the initial latent embeddings: the method refines existing feature geometry but does not generate discriminative information absent from the original representation, so weak or modality-mismatched backbone features limit the benefit of manifold refinement, as seen on datasets with challenging textures or less structured anatomical cues. The framework should thus be viewed as a geometry-enhancement mechanism that complements, rather than replaces, advances in representation learning.

Second, inference is transductive: scoring a test sample requires recomputing empirical density weights and manifold shifts jointly over the combined train-test population (Eq.~\eqref{eq:joint_shift}, stage~(iv)) rather than via a fixed, pre-frozen decision function. This is straightforward in a benchmark setting, where the full test set is available upfront, but in deployment it means each new case must be shifted jointly with a retained reference population (at minimum the training embeddings, in practice also previously screened cases) rather than scored in isolation a departure from fully inductive OCC. While this remains computationally lightweight given the framework's overall efficiency (Section~\ref{sec:results}, Computational Efficiency and Scalability), approximating the transductive shift with a fixed-population proxy or a periodically refreshed reference batch is left for future work.


\section{Conclusion}

In this work, we utilized \textit{Mean Shift-based Feature Refinement} to develop a likelihood-enhancing framework for OCC-based anomaly detection in medical imaging. By combining pretrained feature representations with density-based scoring, the proposed framework addresses a common limitation of conventional pipelines that treat latent feature spaces as fixed after feature extraction. Through an iterative neighborhood refinement process, the method shifts samples toward regions of higher local likelihood, producing feature distributions that are more amenable to Gaussian modeling and Mahalanobis distance-based anomaly scoring, while improving the separability of normal and anomalous samples in feature space without introducing any additional trainable parameters.

Experiments across seven medical imaging datasets spanning five imaging modalities within the MedIAnomaly benchmark~\cite{CAI2025103500} indicate that the proposed framework achieves competitive and frequently state-of-the-art performance, obtaining the highest AUC on four datasets and the highest Average Precision (AP) on five datasets. In particular, the method demonstrated strong joint ranking and precision performance on the Brain Tumor benchmark, where both AUC and AP reached 0.981. In addition, sensitivity analyses using fixed hyperparameters across datasets suggest that the framework remains practically robust without requiring extensive dataset-specific tuning.

From a computational perspective, the proposed framework operates on pre-extracted embeddings and relies primarily on approximate neighborhood search and lightweight iterative updates, making it substantially less demanding than reconstruction-heavy or diffusion-based approaches that require repeated image-space optimization or sampling. Because it operates entirely on latent embeddings rather than raw images, the framework requires no retraining or architectural modification to the upstream feature extractor, making it directly applicable across imaging modalities and an attractive choice for modality-agnostic, scalable deployment settings where efficiency is important.

As noted above, performance remains bounded by the quality of the underlying feature extractor, with ISIC 2018 dermoscopy the clearest example of this dependency accordingly, the framework is best viewed as complementary to, rather than a substitute for, strong representation learning.

Future work will focus on integrating our method within end-to-end trainable pipelines, extending the framework to volumetric imaging modalities such as CT and MRI, exploring domain adaptation across institutions and scanners, and incorporating uncertainty estimation to support clinically reliable anomaly detection systems.

\subsection*{Supplementary information}
To ensure full reproducibility of our results, we make the complete implementation, including all experimental pipelines and configurations, publicly available at: 
\url{https://github.com/D6nam853/medi-msde}.

\section{Statements and Declarations}

\subsection*{Funding}
This research did not receive any specific grant from funding agencies in the public, commercial, or not-for-profit sectors.

\subsection*{Competing interests}
The authors have no competing interests to declare that are relevant to the content of this article.

\subsection*{Data Availability}
The datasets used in this study are publicly available and were accessed through the benchmark repository accompanying the MedIAnomaly study \cite{CAI2025103500}. We follow the standardized dataset splits and preprocessing protocols provided in the official repository  (https://github.com/caiyu6666/MedIAnomaly), which includes RSNA Pneumonia Detection, VinDr-CXR, Brain Tumor Dataset, LAG, ISIC 2018, Camelyon16, and BraTS2021 datasets, subject to their respective licenses and usage terms.

\subsection*{AI-assisted manuscript preparation}

During the preparation of this work, the authors used generative AI tools to assist with code organization, language refinement, and structuring of the manuscript. All generated content was carefully reviewed, edited, and validated by the authors, who take full responsibility for the final content.

\biboptions{numbers,sort&compress}

\bibliographystyle{cas-model2-names}
\bibliography{cas-refs}

\appendix

\section{Hyperparameter Tuning and Model Robustness}
\label{hyperparam}

To rigorously evaluate the sensitivity of the Mean Shift-based Feature Refinement framework to its hyperparameters, we conducted an extensive optimization study using the Optuna framework~\cite{akiba2019optuna}. The primary objective of this analysis was to determine whether the proposed framework relies on brittle, dataset-specific tuning to achieve competitive performance, or if its manifold refinement process is inherently robust across diverse medical imaging modalities.

\paragraph{Zero-Leakage Validation Protocol.}
Hyperparameter optimization in anomaly detection carries a high risk of data leakage, particularly when anomalous samples are scarce. To prevent this, we implemented a strict "zero-leakage" validation protocol. For each dataset, we constructed an isolated validation set consisting of:
\begin{enumerate}
    \item \textbf{Validation Normals:} 20\% of the normal training samples were held out.
    \item \textbf{Validation Anomalies:} 10\% of the anomalous samples were permanently removed from the final test set and repurposed strictly for validation. 
\end{enumerate}
During the Optuna search, our model was fitted exclusively on the remaining 80\% of the normal training data and evaluated on the validation set. The final test set remained completely unseen during the entire tuning process.\\

\paragraph{Search Space and Optimization.}
The optimization process focused strictly on the five geometric hyperparameters governing the manifold shift process: the number of neighbors ($k \in [5, 60]$), neighborhood sample density threshold ($\tau \in [3, 80]$), learning rate ($\eta \in [0.01, 0.5]$), maximum shift iterations ($T \in [3, 12]$), and the shift convergence threshold ($\delta \in [10^{-4}, 0.05]$). The Gaussian Density Estimation (GDE) parameters, including the PCA dimensionality reduction ($d'=256$), were kept fixed. The objective metric was set to maximize the Area Under the ROC Curve (AUC) over 80 trials per dataset.

\paragraph{Results and Robustness Analysis.}
Following the Optuna search, the model was retrained on the full set of normal training samples using the dataset-specific optimal hyperparameters and evaluated on the held-out test set. 

Table~\ref{tab:optuna_robustness} shows a mixed but informative picture. On four of the seven datasets (RSNA, Brain Tumor, LAG, BraTS2021), Optuna-tuned hyperparameters match or modestly improve upon the universal defaults, by at most 1-2 AUC/AP points. On the remaining three datasets, however, tuning under our zero-leakage validation protocol actually \emph{degrades} test performance relative to the fixed defaults: VinDr-CXR drops by 5.8 AUC / 6.1 AP points, Camelyon16 by 2.7 AUC / 8.1 AP points, and ISIC~2018 by 1.0 AUC / 6.0 AP points.

We interpret this not as evidence that the framework is insensitive to its hyperparameters in an absolute sense, but as evidence that, under a realistic, leakage-free validation budget, small and noisy validation sets can yield hyperparameter choices that overfit to validation idiosyncrasies rather than to genuine dataset structure. In this light, the universal default configuration is not merely a convenient simplification but a \emph{safer} choice in practice: it avoids the risk of validation-set overfitting inherent to per-dataset tuning when abnormal samples are scarce, while still achieving state-of-the-art or near-state-of-the-art performance across all seven datasets (Table~\ref{tab:benchmark_results}). This supports the practical recommendation of deploying the framework with fixed hyperparameters in low-label clinical settings, rather than the stronger claim that performance is uniformly insensitive to hyperparameter choice.

\begin{table*}[t]
\centering
\caption{Comparison of our model's performance using universal (fixed) hyperparameters versus Optuna-optimized, dataset-specific hyperparameters under the zero-leakage validation protocol. Tuning yields small gains on four datasets (RSNA, Brain Tumor, LAG, BraTS2021) but underperforms the fixed defaults on the remaining three (VinDr-CXR, Camelyon16, ISIC~2018), supporting the use of universal hyperparameters as a safer default when validation data is limited. Best results per metric are highlighted in \textbf{bold}.}
\label{tab:optuna_robustness}
\resizebox{\textwidth}{!}{%
\begin{tabular}{l cc cc l}
\toprule
\textbf{Dataset} & \multicolumn{2}{c}{\textbf{Universal (Fixed)}} & \multicolumn{2}{c}{\textbf{Optuna-Tuned}} & \textbf{Optimal Tuned Hyperparameters} \\
\cmidrule(lr){2-3} \cmidrule(lr){4-5}
& \textbf{AUC} & \textbf{AP} & \textbf{AUC} & \textbf{AP} & $k$, $\tau$, $\eta$, $T$, $\delta$ \\
\midrule
RSNA Pneumonia & 0.918 & 0.906 & \textbf{0.923} & \textbf{0.908} & $k$=10, $\tau$=32, $\eta$=0.16, $T$=9, $\delta$=0.0067 \\
VinDr-CXR & \textbf{0.819} & \textbf{0.797} & 0.761 & 0.736 & $k$=50, $\tau$=18, $\eta$=0.14, $T$=8, $\delta$=0.0001 \\
Brain Tumor & 0.981 & 0.981 & \textbf{0.991} & \textbf{0.990} & $k$=25, $\tau$=55, $\eta$=0.26, $T$=6, $\delta$=0.0001\\
LAG & 0.810 & \textbf{0.831} & \textbf{0.817} & 0.816 & $k$=32, $\tau$=79, $\eta$=0.43, $T$=4, $\delta$=0.0003 \\
BraTS2021 & 0.736 & \textbf{0.867} & \textbf{0.752} & 0.848 & $k$=45, $\tau$=41, $\eta$=0.01, $T$=3, $\delta$=0.0049 \\
Camelyon16 & \textbf{0.812} & \textbf{0.820} & 0.785 & 0.739 & $k$=15, $\tau$=76, $\eta$=0.47, $T$=3, $\delta$=0.0001 \\
ISIC 2018 & \textbf{0.705} & \textbf{0.638} & 0.695 & 0.578 & $k$=18, $\tau$=63, $\eta$=0.19, $T$=11, $\delta$=0.0110 \\
\bottomrule
\multicolumn{6}{l}{\footnotesize \textit{Note:} Universal hyperparameter values are fixed at $k$=50, $\tau$=70, $\eta$=0.33, $T$=8, $\delta$=0.01 across all datasets.} \\
\end{tabular}%
}
\end{table*}

\section{Details on GradCAM implementation}
\label{gradcam}
We employ Gradient-weighted Class Activation Mapping (GradCAM)~\cite{selvaraju2017gradcam} to produce saliency heatmaps by backpropagating the Mahalanobis anomaly score through the frozen backbone to its final convolutional layer (\texttt{layer4} for ResNet18-based architectures). Crucially, this analysis is conducted entirely within the one-class classification (OCC) paradigm: the backbone is frozen throughout, the reference Gaussian ($\boldsymbol{\mu}_z$, $\boldsymbol{\Sigma}_z^{-1}$) and PCA projection are fitted exclusively on the shifted normal training embeddings, and no anomaly labels are used in the saliency computation. Ground-truth labels are used solely for selecting representative abnormal test images for visualization.

For a given test image, the forward pass through the backbone produces both a spatial feature map tensor $\mathbf{A} \in \mathbb{R}^{512 \times 7 \times 7}$ (prior to global average pooling) and the resulting 512-dimensional embedding. The Mahalanobis score is then computed using the Gaussian parameters and PCA projection fitted on the manifold-shifted training distribution. Because the manifold shift is a non-differentiable operation (implemented via NNDescent and Numba JIT on pre-extracted embeddings), it is not part of the gradient path. Instead, we re-implement the scoring pipeline StandardScaler normalization, PCA projection, and Mahalanobis distance as a fully differentiable sequence of linear operations in PyTorch, using the parameters already fitted during the models training phase. Since the manifold shift acts after global average pooling, it does not alter the spatial structure of the feature maps; it only rescales the resulting gradient uniformly, leaving the spatial saliency pattern intact. The gradient $\partial s / \partial \mathbf{A}$ therefore faithfully reflects which spatial regions drive the anomaly score under the full framework. The GradCAM heatmap is obtained as:
\begin{equation}
    \mathcal{M}_{\text{GradCAM}} = \text{ReLU}\!\left(\sum_{c} \alpha_c \cdot 
    \mathbf{A}^c\right), \quad 
    \alpha_c = \frac{1}{H \times W}\sum_{i,j}\frac{\partial s}{\partial \mathbf{A}_{ij}^c},
\end{equation}
where $\alpha_c$ is the importance weight for channel $c$ and $\mathbf{A}^c$ is the $c$-th channel of the feature map. The resulting heatmap is upsampled to the input resolution and overlaid on the original image.

\end{document}